\documentclass[conference]{IEEEtran}
\IEEEoverridecommandlockouts
\usepackage{cite}
\usepackage{amsmath,amssymb,amsfonts}
\usepackage{algorithmic}
\usepackage{graphicx}
\usepackage{textcomp}
\usepackage{xcolor}
\usepackage{subcaption}
\def\BibTeX{{\rm B\kern-.05em{\sc i\kern-.025em b}\kern-.08em
    T\kern-.1667em\lower.7ex\hbox{E}\kern-.125emX}}
\begin{document}

\title{A Comparative Analysis of Registration Tools: Traditional vs Deep Learning Approach on High Resolution Tissue Cleared Data
\thanks{A. Nazib is supported by a QUT Postgraduate Research Award (QUTPRA).}
}

\author{\IEEEauthorblockN{Abdullah Nazib, Clinton Fookes, Dimitri Perrin}
\IEEEauthorblockA{\textit{School of Electrical Engineering and Computer Science} \\
\textit{Queensland University of Technology}\\
Brisbane, Australia \\
abdullah.nazib@hdr.qut.edu.au}
}



\maketitle

\begin{abstract}
Image registration plays an important role in comparing images. It is particularly important in analyzing medical images like CT, MRI, PET, etc. to quantify different biological samples, to monitor disease progression and to fuse different modalities to support better diagnosis. The recent emergence of tissue clearing protocols enable us to take images at cellular level resolution. Image registration tools developed for other modalities are currently unable to manage images of such extreme high resolution. The recent popularity of deep learning based methods in the computer vision community justifies a rigorous investigation of deep-learning based methods on tissue cleared images along with their traditional counterparts. In this paper, we investigate and compare the performance of a deep learning based registration method with traditional optimization based methods on samples from tissue-clearing methods. 
From the comparative results it is found that a deep-learning based method outperforms all traditional registration tools in terms of registration time and has achieved promising registration accuracy. 
\end{abstract}

\begin{IEEEkeywords}
Tissue Clearing, Image Registration, Optimization
\end{IEEEkeywords}

\section{Introduction}
Tissue clearing is a complex bio-chemical process that can remove light-obstructing biological elements from tissues. Clearing of obstructive elements enable us to take images of cellular level details like the connection between cells. Comparison across multiple samples can also reveal the structure and dynamics of cellular activities. Images obtained from tissue clearing are very high in resolution and extremely small in pixel spacing. A simple mouse brain sample from tissue clearing can provide a 3D image of around 6 gigabytes, 1000 times larger than a typical human brain MRI image. In terms of resolution, a tissue cleared sample (from the CUBIC dataset used in this study \cite{Susaki2014}) has a voxel resolution in the micro-meter scale ($6.45\times 6.45\times10\mu m^3$) whereas a MRI of a human brain usually has a resolution in the millimeter scale ($0.86 \times 0.86 \times 1.5 mm^3$). Therefore, the computational analysis of these images is challenging. \\
Image registration is a computational method that is used in image analysis and is an essential step in the comparison of 
images. Despite having a long history of research and development of medical image
registration, most state-of-the-art registration algorithms are optimization based and do complex computation on image data iteratively until a satisfactory convergence condition is achieved. As the volume of data increases the computation time increases dramatically. For downscaled tissue cleared images, the best performing image registration tool ANTS \cite{AVANTS2008} takes more than 8 hours to register a pair of brain samples \cite{Nazib2018}. This is a serious bottleneck and raises questions regarding their applicability to full-scale data.
Deep learning has become popular in computer vision and has achieved very promising performance in tasks such as object detection and natural language processing. In medical imaging, deep learning is being used extensively in organ segmentation \cite{yu2017, dou2017} and performs with high precision. Due to the complex nature of image registration and the lack of ground-truth registration data, the direct application of deep learning based models in image registration is not straight forward.
\\In the literature \cite{Litjens2017}, two different approaches of deep learning based image registration methods are found. In one approach, CNN architectures are used to estimate the similarity measurement between moving and fixed images and the CNN architecture is integrated in an optimization framework. Methods like \cite{Simonovsky2016}, \cite{Cheng2016} and  \cite{Wu2016} falls in this approach. 
In the second approach, deep-architectures are directly applied on image data to estimate transformation parameters.
Approaches like \cite{YangKN16}, fall into this category. Methods discussed here are all applied on MRI data. 
\\In our previous work \cite{Nazib2018}, we performed an evaluation of optimization based tools using tissue cleared images. In this paper, we extend our evaluation with five publicly available optimization based image registration tools (ANTS, AIR, Elastix, NiftyReg, IRTK) and an unsupervised deep-learning based registration method (VoxelMorph) to perform registration tasks on tissue cleared images from the CUBIC \cite{Susaki2014} dataset. We compare the performance of the selected tools qualitatively and quantitatively along with their efficiency in time.    

\section{Registration Tools Setup}
\subsection{Optimization Based Methods}
1) \textbf{IRTK:} One of early image registration tool for breast MRI images using voxelized mutual information similarity and free-form deformation model \cite{Rueckert1999} . Before starting registration IRTK apply contrast enhancement to make similarity measure insensitive to intensity change. A hierarchical transformation model is applied to capture global and local motion of the volume data where global motion captured by affine model and local motion is by non-linear free form deformation model. Voxel-based Normalized mutual information is used as the similarity measure.   
In this evaluation, we followed the same settings as \cite{Xu2016} except B-spline control points. Since the pixel spacing of CUBIC dataset is very small, the control point spacing is set to 5mm which is the highest possible value for this method.        
\\
2) \textbf{AIR:}The earliest automatic registration tool \cite{Woods1993}. In AIR parameter settings we applied defaults threshold values for rigid and affine transformation given in their documentation. Using affine registered parameters as the starting points for non-linear registration, second order non-linear registration is performed and then 3rd order non-linear registration is initialized and applied which is the maximum possible registration for CUBIC dataset.
\\
3) \textbf{Elastix:} One of the popular registration tool developed for CT and MRI images with large set of common registration algorithms \cite{Klein2010}. This tool consists of many algorithms for similarity, optimization, regularization, interpolation, transformation etc. For similarity measure Elastix include mutual information (MI), Normalized mutual information (NMI), Cross-correlation (CC), mean squared difference (MSD) etc. The transformation model included in Elastix library are: rigid, affine with different degree of freedom, B-spline with physics based control points in uniform and non-uniform grids. And a set of optimization methods namely gradient descent, quasi-Newton, nonlinear conjugate gradient (with several variants), and a number of stochastic gradient descent methods. All these options adds flexibility to choose required components whenever necessary.    
In this evaluation, we consider elastix parameter settings used in \cite{Hammelrath2016} for rigid, affine and nonlinear registration. 
\\
4) \textbf{ANTS:} Advanced Normalization Tool (ANTS) use symmetric diffeomorphic normalization method for non-linear transformation \cite{AVANTS2008, Avants2011}. In ANTS, cross-correlation is maximized in symmetric diffeomorphic map and use Eular-Lagrange equations for optimization. The diffeomorphic map preserves topology map along with invertible transformation parameters and gives sub-pixel accuracy.    
The parameter settings for ANTS tool is derived from ANTS example script. In evaluation by \cite{Xu2016}, two different setup of ANTS tool were used with two different similarity metric (Cross-Correlation and Mutual Information) which they considered as two separate methods. In our settings, we used cross-correlation as similarity measure, the number of resolution level is 3 with 100 iteration in each sampling level.  
\\
5) \textbf{NiftyReg:} It is also a promising registration tool \cite{Modat2010}. It is an extended version of IRTK, based on free form deformation. In this method, the gradient of normalized mutual information of each B-spline control points are calculated and used in gradient descent based optimization method. The algorithm was implemented with parallel processing, but in our evaluation only CPU version is used. The parameter setting of this tool for CUBIC evaluation is exactly same as settings mentioned by \cite{Xu2016}. The number of iteration 1000 for free-form deformation and 500 intensity threshold for both source and target image.

\subsection{Deep-Learning Based Method}
A number of deep learning based image registration methods has been developed \cite{Rohe},\cite{Li2018},\cite{Sokooti},\cite{YangKN16}. Among all of these methods, we select Voxlemorph\cite{Balakrishnan2018} method due to its high registration accuracy and ease of training methodology.\\
\textbf{VoxelMorph:} This is an unsupervised deep-learning based registration method \cite{Balakrishnan2018}. It integrates a fully convolutional U-net architecture with a Spatial transformer and train them simultaneously. Unlike other deep-learning based approaches, this architecture directly takes fixed and moving images in 3D form instead of taking image patches. Using cross-correlation as objective function to estimate dissimilarity between fixed and instantaneously moved moving image, the network is trained in unsupervised manner. A diffusion regularization is used to prevent training over-fitting of the network.
\section{Data Preprocessing and Generation}
In this work, tissue cleared images of Arc-dVenus mouse brains are used for both types of registration methods. The process of clearing tissue of mouse brain is a completely biochemical process and out of scope of this work. Following is the brief discussion of the process involved in data acquisition and data pre-processing:      
\subsection{Tissue Clearing}
There are many tissue clearing methods have been developed. Among them, popular methods include BABB\cite{Dodt2007}, Scale \cite{Hama2011}, SeeDB \cite{Ke2013}, CLARITY \cite{Chung2013} and iDISCO \cite{Renier2014}. In this paper we used data from a tissue clearing method named CUBIC protocol. In this method data is acquired in three steps: tissue clearing by chemical process, image acquisition,
and image analysis (including registration). The chemical process can be performed in three different ways: i) Simple immersion protocol for
whole-organ clearing, ii) CB-perfusion and immersion protocol for faster clearing for whole
organs, and iii) CB-perfusion protocol for whole-body clearing \cite{Susaki2014,Ke2013}. Based on the protocols
applied, it takes 11 to 14 days to clear an organ or whole body.

\subsection{Data Acquisition}
The imaging process is detailed in \cite{Susaki2015} and summarized here for convenience. After bio-chemical process of tissue clearing, a Light-Sheet Fluorescence Microscope (LSFM) with CMOS camera of
resolution $2560\times2160$ pixels and a pair of laser excitation and emission filters of wavelength 488nm
for green signal and 588nm for red signal are used for rapid 3D imaging. The acquired
images have resolution of $6.5\times6.5$ micrometer, which is sufficient to detect signals from a
single cell. For whole brain sample with clearer and sharper 3D image volume, images of the mouse brain are taken in two
opposite directions, Dorstal-to-Ventral (D-V) and Ventral-to-Dorsal (V-D) respectively from same brain. 
Thus the LSFM imaging procedure takes four 2D stacks for each brain: two channels (one
for cell positioning and one for cell activity), and two directions.

\subsection{Data Preparation for Optimization based methods}
The obtained V-D and D-V stacks, mentioned in previous section, are registered for alignment 
(using the cell positioning data) and then merged together. Before this alignment, the data has to be downscaled due to software limitation (no existing software tool can handle the high-resolution images of 100\% size).
The merging of two stacks is done by calculating edge content, and this edge information is used to
determine which slices should be taken from which imaging direction. For each brain sample this procedure generates 
one structure volume and one signal volume in Nifti format. Only structural volumes are used in this evaluation. 
For the purpose of this evaluation, we downscale 100\% resolution files to 10\% and 15\% using the procedure explained in the published study \cite{Tatsuki2016}. The resolution downscaling is done by python code from CUBIC protocol \cite{Susaki2015}.
In this study, data from \cite{Susaki2015} is used. In total we obtain 23 different CUBIC brains from this clearing protocol. For testing traditional methods, we use data from \cite{Susaki2015}. We select brain ``003" as reference and registered other two brains to it in all resolution level.

\subsection{Data Generation for Deep-Learning Method}
To train the selected Voxelmorph network, a large training dataset is required. To train and test VM architecture we generate a large synthetic dataset from CUBIC dataset. To create a large training dataset, we used 16 CUBIC brains from separate dataset. For these 16 brains, a flipping operation is applied in X,Y,Z direction and then XYZ direction altogether. Thus, a set of 80 brains are generated including the original 16 CUBIC brains. For each of these brains we apply 100 randomly generated deformation fields and obtain 8000 brains in total as training dataset. To generate deformation field, 150 random locations are selected from a 3D brain. To make these selected locations as center of smooth folding, Gaussian blurring with three different frequency (25, 35, 45) is applied. Three frequency of Gaussian blurring is used to make the artificially generated brains more realistic and consistent with original CUBIC brains. Also, these different blurring frequencies give different versions of smoothing on deformation fields and ensure variability among them. In this way we generate 100 deformed version of each brain where 20 brains are in low frequency (25), 20 brains are in medium frequency (35) and 60 are in high frequency (45). We make brain ``003" of first batch as reference brain(moving image) and take all other generated brains as Fixed image and train the network.

\begin{figure}[t]
	\begin{center}
		\includegraphics[trim=10cm 10cm 10cm 10cm,scale=0.45]{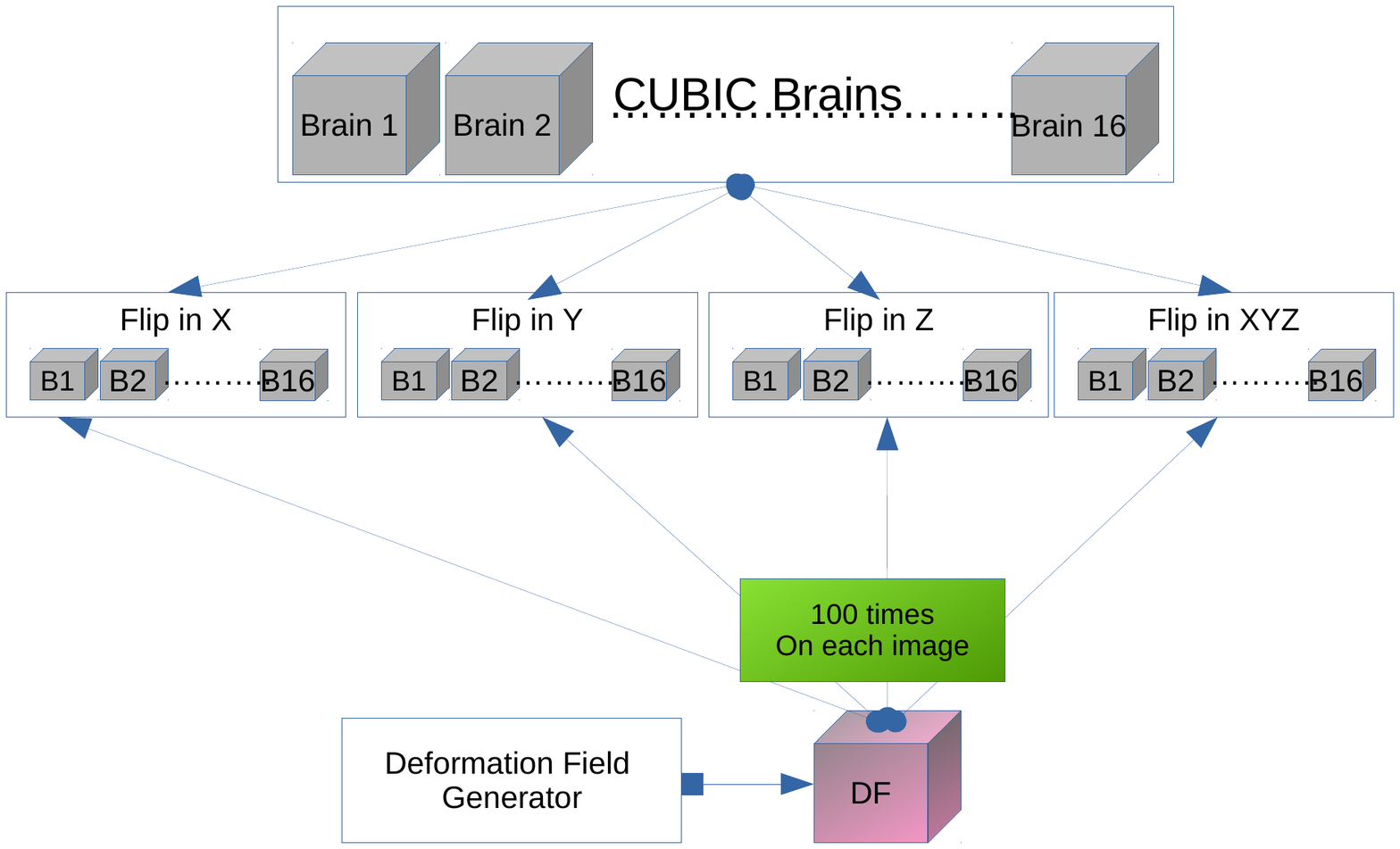}
	\end{center}
	\caption{Data Generation Process for the Training Set}
	\label{Figure:1}	
\end{figure}

\subsection{Experimental Setup for Optimization Based Methods}
Since traditional methods do not require training, we use only the first batch for evaluation. We use brain ``003" as reference brain and registered other two brains against it. Quantitative and qualitative comparison of selected tools are measured. We run the experiment in a high performance computing environment with 2.66GHz 64 bit Intel Xeon processor using 256GB memory.  

\subsection{Experimental Setup for Deep-learning Based Method}
Similar to optimization based methods, the deep learning based VoxelMorph is also trained and tested on HPC environment. Since network is developed using tensorflow, we ran the training on High Performance Computing (HPC) environment with GPU of 12GB video memory and additional 64GB RAM. We train the network using artificially deformed 
brains and making the brain "003" as reference brain. In each training iteration, the network randomly picks one generated brain as fixed image and picks brain ``003" as reference. 

\section{Results}
\subsection{Quantitative Evaluation}
As mentioned earlier, brain sample "003" is used as reference brain and brain samples 1 and 2 are aligned to this reference. For two different resolution level of CUBIC brains, we run the experiment with same settings and parameters. The results are noted in Table \ref{Table:10} and Table \ref{Table:15} for 10 and 15\% resolution respectively. 
Both table shows the cross-correlation and mutual information measures of each registration tool. In Table \ref{Table:10}, for both evaluation metric and brain, ANTS obtained the best scores with greater than 0.9. Elastix obtained second best CC scores with 0.95 for brain 1 and 0.93 for brain 2. With MI, Elastix suppresses ANTS with 3.319 for brain 1 while it is slightly smaller for brain 2. NiftyReg obtained third best with scores in the range 0.7 to 0.87 in both metric. Both IRTK and AIR scored under 0.7 in all cases except IRTK on brain 1. The deep-learning based Voxelmorph scored  0.89 and 0.91 in CC with around 120,000 (288 hours approximately) iteration which is between Elastix and NiftyReg. To improve the results, we run the network with more iterations around 200,000 and obtain performance better than NiftyReg and slightly lower than Elastix. We use default settings for both iterations. Training this heavy network with large number of iteration requires long walltime(500 hours approximately). To reduce training time and to make the network more consistent on CUBIC dataset, network parameters are tuned. The Voxelmorph cost function has a weighting parameter $\lambda$ that controls the regularization in the cost function. In the original paper \cite{Balakrishnan2018} of Voxelmorph, performance of network with different settings of parameter $\lambda$ is presented. In default setting $\lambda=1.0$. For CUBIC data, we set $\lambda=1.5$ and take performance measures at different number of iteration. With this tuned setting, it is found that with only 10000 iteration performance reaches to the level of 120,000 iteration of default settings. We continue training the network up to 80,000 (192 hours approximately) iteration and performance improved slightly but very slowly.                 

\begin{table}[t]
	\begin{center}
		\caption{CC and MI Score at 10\% Resolution}
		\label{Table:10}
		\begin{tabular}{l c c c c c}
		\hline
		Methods		& iteration & brain 1 & brain 1 & brain 2 & brain 2\\
				&  & CC & MI & CC & MI\\
		\hline
		ANTS		&100				&0.9494		&0.9127			&0.9596		&0.9372\\
		AIR			&--					&0.5138		&0.7437			&0.4841		&0.7690\\
		Elastix		&1000				&0.9589		&3.3194			&0.9388		&0.8612\\
		NiftyReg	&1000				&0.8623		&0.8009			&0.8489		&0.7945\\
		IRTK		&1000				&0.8013		&0.5872			&0.6958		&0.4593\\
        VM			&119800				&0.8962 	&0.3401			&0.9143 	&0.3578\\
		VM			&199950				&0.9063		&0.3473			&0.9204 	&0.3594\\
		VM tuned 	&10000				&0.8364		&0.4394			&0.8850		&0.4758\\
        VM tuned 	&20000				&0.8523		&0.4570			&0.8986		&0.4938\\
        VM tuned 	&30000				&0.8555		&0.4612			&0.9049		&0.5078\\
		VM tuned 	&40000				&0.8555		&0.4612			&0.9057		&0.5031\\
        VM tuned 	&50000				&0.8592		&0.4614			&0.9056		&0.5057\\
		VM tuned	&80000				&0.8719		&0.4686			&0.9098		&0.5082\\			
		\hline
		\end{tabular}
	\end{center}
\end{table}

For 15\% resolution, similar pattern of performance measure is found. ANTS, Elastix and NiftyReg remain the top scorers among the traditional registration methods. The deep-learning based method also shows similar pattern with gradual increase of performance in all brains and measurements. The only notable difference is at 10\% resolution; MI scores are low compared to MI scores at 15\% resolution. 

\begin{table}[t]
	\begin{center}
		\caption{CC and MI Score at 15\% Resolution}
		\label{Table:15}
		\begin{tabular}{l c c c c c}
		\hline
		Methods		& iteration & brain 1 & brain 1 & brain 2 & brain 2\\
					&  & CC & MI & CC & MI\\
		\hline
		ANTS &100				&0.9386		&1.7889			&0.9530		&1.8170\\
		AIR		&--					&0.3000		&1.3662			&0.3290		&1.4066\\
		Elastix	&1000				&0.9161		&1.7218			&0.9261		&1.8068\\
		NiftyReg 	&1000				&0.6802		&2.3915			&0.7693		&1.8518\\
		IRTK		&1000				&0.0003		&0.0359			&0.0023		&0.0359\\
		VM tuned 	&5000 		&0.6857		&0.7411			&0.6395		&0.7296\\
		VM tuned 	&10000		&0.7175		&0.7749			&0.6720		&0.7641\\
                VM tuned 	&20000		&0.7532		&0.8050			&0.6922		&0.7703\\
                VM tuned 	&30000		&0.7581		&0.7959			&0.6864		&0.7383\\
		VM tuned 	&40000		&0.7839		&0.8216			&0.7029	  	&0.7615\\
		VM tuned 	&49340		&0.7853		&0.8226			&0.7114		&0.7647\\			
		\hline
		\end{tabular}
	\end{center}
\end{table}

With tuned parameter settings at 10\%, it is found that after 20,000 iterations, accuracy improvement in both metrics for both brains is very small. For 15\% resolution, we train the network in HPC with same settings and check the accuracy on different iterations. For 15\%, at 5000 (22 hours approximately) iteration, CC is only 0.68 and 0.64 while MI is 0.74 and 0.73 respectively. At 20,000 iteration, accuracy improvement is small and at 30000 iteration its even smaller. A similar pattern is found between 40,000 (176 hours approximately) and near 50000 (220 hours approximately) iterations. With tuned parameter settings, network learns more quickly than default settings. More importantly, with the tuned settings, obtained mutual information scores are improved slowly along with cross-correlation scores. 

\subsection{Qualitative Evaluation}
\begin{figure}[h!]
	\begin{center}
    	\begin{minipage}[t]{4.5cm}
    		\includegraphics[width=4.3cm,height=4.3cm]{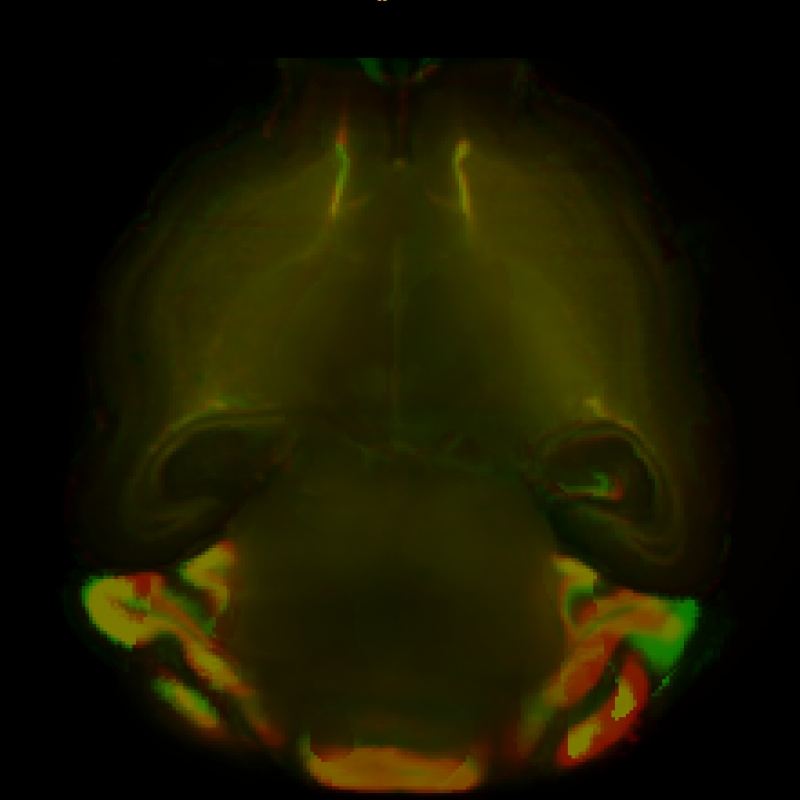}
    		\subcaption{ANTS 10\%}
    		\label{fig:InterVis:ANTS10}
    	\end{minipage}%
    	\begin{minipage}[t]{4.5cm}
    		\includegraphics[width=4.3cm,height=4.3cm]{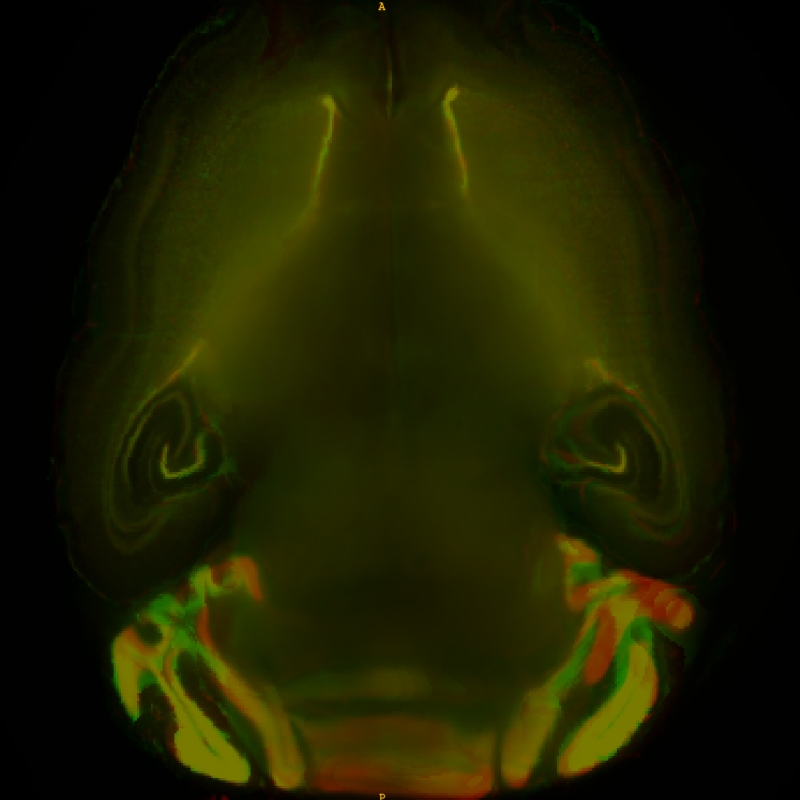}
    		\subcaption{ANTS 15\%}
    		\label{fig:InterVis:ANTS15}
    	\end{minipage}%
    	\vfill
    	\begin{minipage}[t]{4.5cm}
    		\includegraphics[width=4.3cm,height=4.3cm]{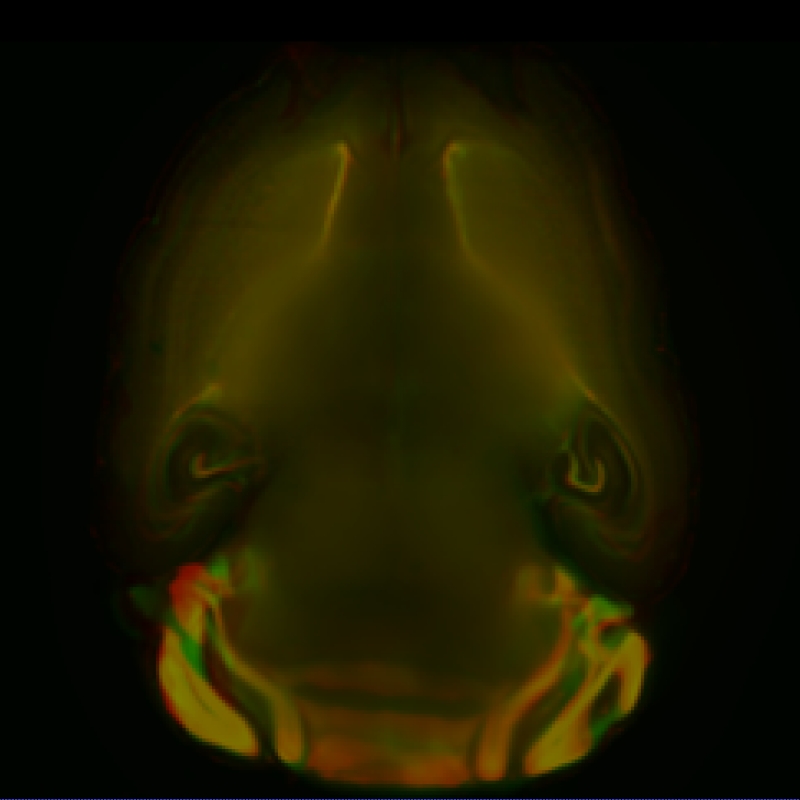}
    		\subcaption{Elastix 10\%}
    		\label{fig:InterVis:Elastix10}
	\end{minipage}%
    	\begin{minipage}[t]{4.5cm}
    		\includegraphics[width=4.3cm,height=4.3cm]{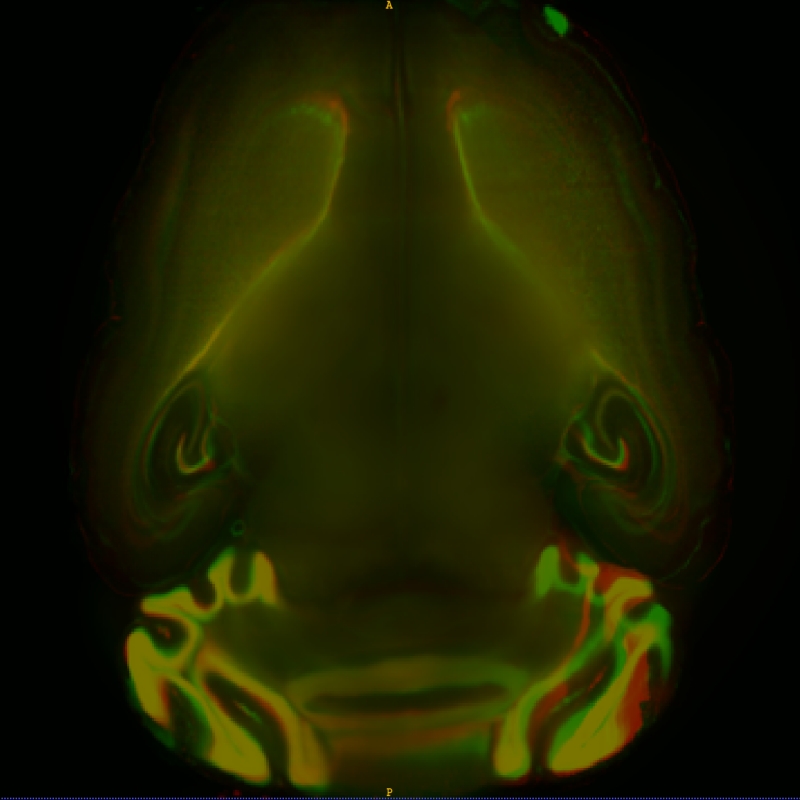}
    		\subcaption{Elastix 15\%}
    		\label{fig:InterVis:Elastix15}
    	\end{minipage}%
    	\vfill
    	\begin{minipage}[t]{4.5cm}
    		\includegraphics[width=4.3cm,height=4.3cm]{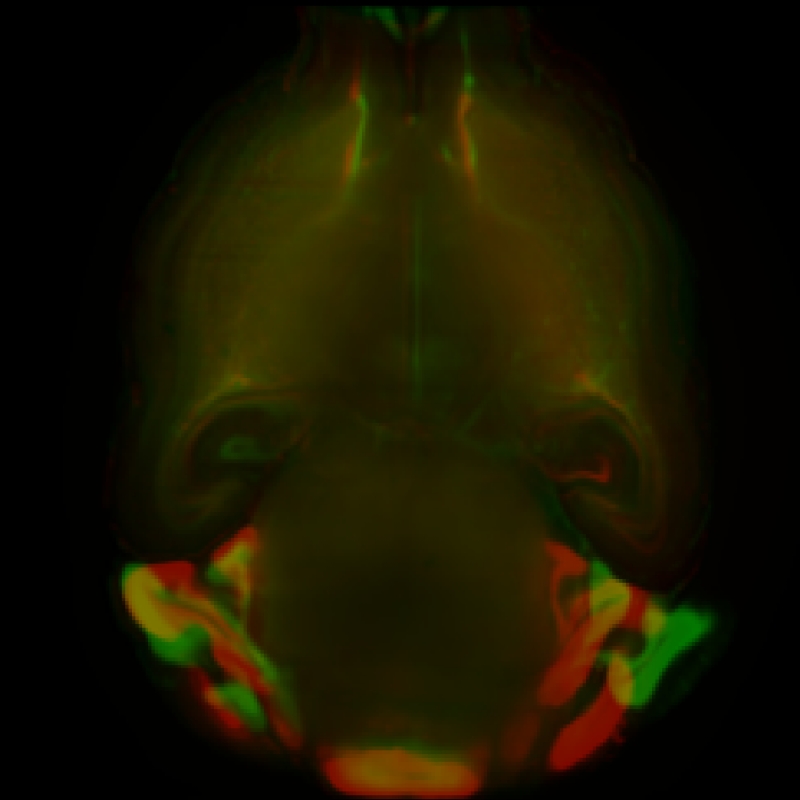}
    		\subcaption{NiftyReg 10\%}
    		\label{fig:InterVis:NiftyReg10}
    	\end{minipage}%
    	\begin{minipage}[t]{4.5cm}
    		\includegraphics[width=4.3cm,height=4.3cm]{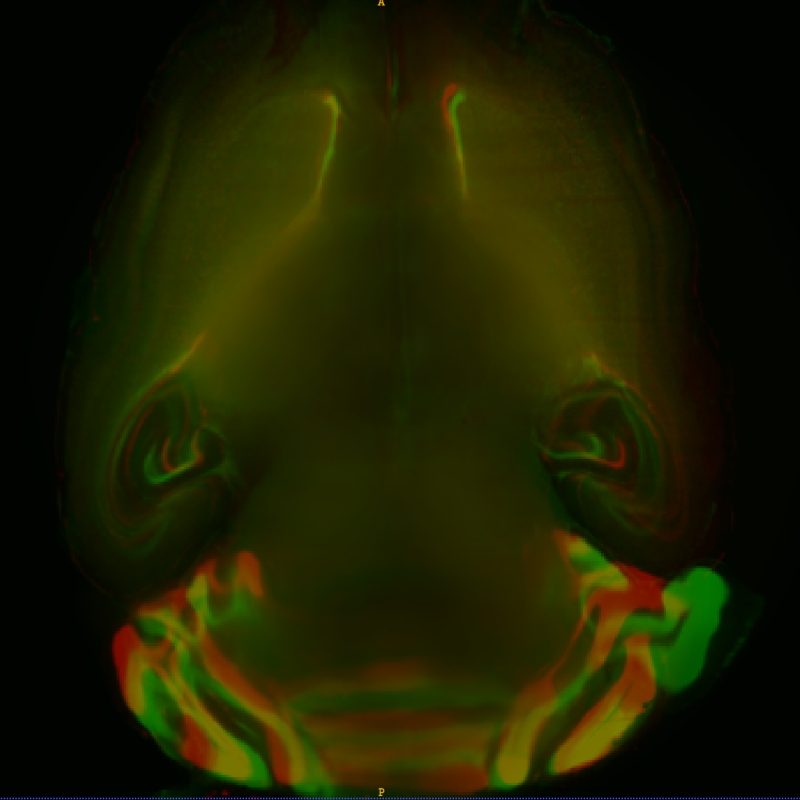}
    		\subcaption{NiftyReg 15\%}
    		\label{fig:InterVis:NiftyReg15}
    	\end{minipage}%
    	\vfill
    	\begin{minipage}[t]{4.5cm}
    		\includegraphics[width=4.3cm,height=4.3cm]{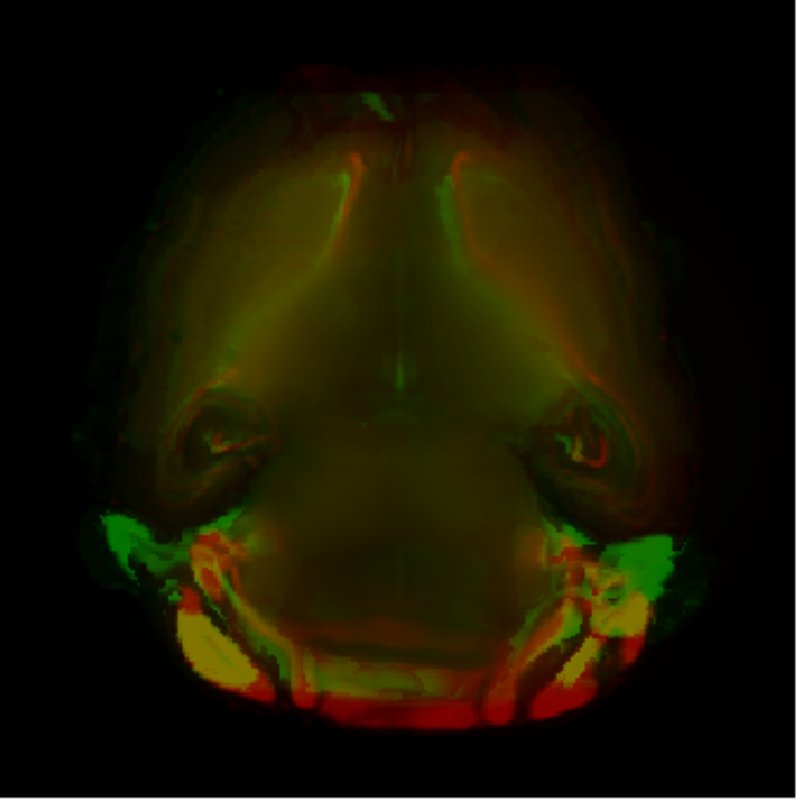}
    		\subcaption{Voxelmorph 10\%}
    		\label{fig:InterVis:Vox10}
    	\end{minipage}%
    	\begin{minipage}[t]{4.5cm}
    		\includegraphics[width=4.3cm,height=4.3cm]{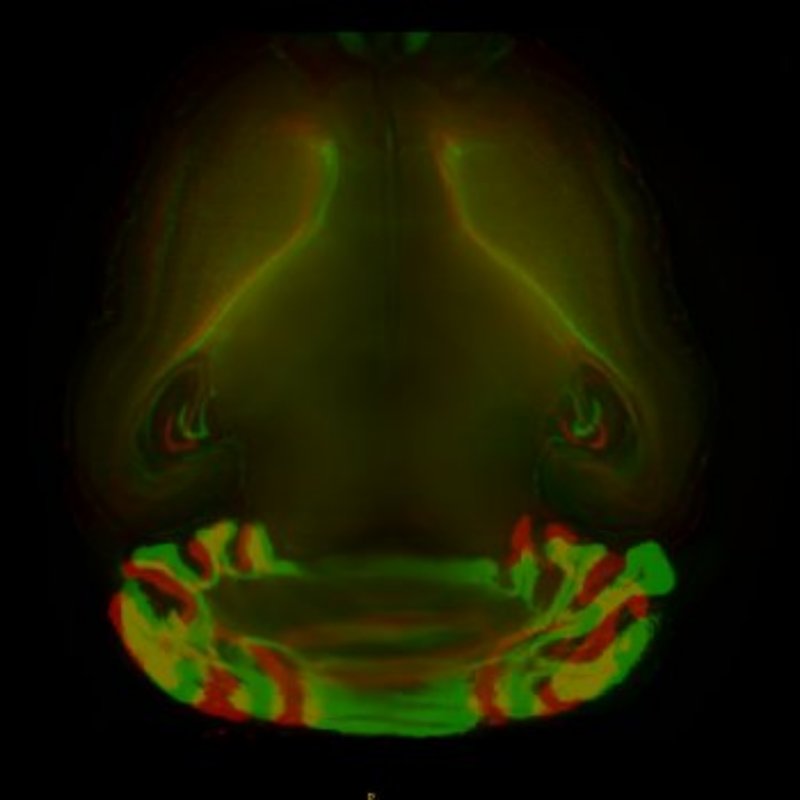}
    		\subcaption{Voxelmorph 15\%}
    		\label{fig:InterVis:Vox15}
    	\end{minipage}%
    	\hfill
	\end{center}
	\caption{Visual Comparison of Registration Tools on 10\% and 15\% resolution}
	\label{fig:qualitative_results}	
\end{figure}

In Figure~\ref{fig:qualitative_results}, the qualitative registration results of top performing optimization based registration algorithms (ANTS, Elastix and NiftyReg) are shown along with deep-learning based Voxelmorph. For the ease of visual analysis, reference brain is shown in red and overlaid with the aligned brain, shown in green. In Figure \ref{Figure:2}, the first row shows registration performance of ANTS on two different resolution level. The visible differences in the cerebellum region of brain is expected for all the methods used. The differences in other regions like hippocampal formation and  dentate gyrus is small and this small difference indicates accuracy of the ANTS registration algorithm.
Good performance of Elastix tool is found in at 10\%, but the performance degrades as the resolution increases. For 15\% resolutions, registration performed by Elastix shows more differing mismatch in hippocampal formation and other region towards cerebral cortex. Similar pattern of misalignment is also visible in NiftyReg registration. 
For Voxelmorph  at 10\% resolution, misalignment is similar to NiftyReg 10\%. The difference in cerebellum region is high and non smooth compared to other methods. Interestingly at 15\% resolution, registered image is more smooth and misalignment between registered image and reference image reduces which is opposite to the Elastix and NiftyReg. Due to the GPU memory limitation, Voxelmorph is unable to operate at higher resolutions. Promising results in these two resolutions is a clear indication of potentials of deep-learning based algorithms for image registration.

\subsection{Efficiency Evaluation}
The efficiency of each registration tool is shown in Table \ref{Performance}.
No direct relationship between accuracy and efficiency is observed. AIR performed very efficiently but, but its accuracy is not satisfactory at all. IRTK is neither good in accuracy nor in efficiency. 
Among the top three performers in optimization based methods, ANTS is the least efficient, with completion times of more than 4 hours for registration at 15\% resolution. Elastix proved to be the most efficient of the three, with both resolution sizes being processed in less than an hour on HPC environment.

Deep-learning based Voxelmorph on the other hand performed extremely well compared to optimization based counter parts. In testing time registration is simply a feed-forward pass which is nothing but a matrix multiplication.It took only 13 seconds at 10\% resolution and approximately 15 seconds at 15\% resolution. This very high efficiency and comparable accuracy, makes it a promising registration tool.  

\begin{table}[t]
	\small
	\begin{center}
		\caption{Computation time of Registration}
		\label{Performance}
		\begin{tabular}{l c c}
			\hline
			Methods & 10\% & 15\%\\
			\hline
			ANTS		&01:23:44   &04:13:18   \\
			AIR			&00:00:31   &00:25:51 	\\
			Elastix 	&00:05:03  	&00:06:30 	\\ 
			NiftyReg	&00:16:32   &00:44:55	\\
			IRTK		&00:02:42   &03:09:26 	\\
            Voxelmorph  &00:00:13	&00:00:16\\
			\hline
		\end{tabular}
	\end{center}
\end{table}
\subsection{Reference Test of Deep-learning Method}
\label{sec:RefenceTest}
To verify the applicability of deep-learning based Voxelmorph as a generic registration algorithm like optimization based algorithms, we decide to test the trained network using a different reference image. During training, we used brain ``003" as reference brain. The trained network is then tested with a different reference brain and the same testing set. 
Table \ref{Table:more_test} shows the results at 15\% resolution. For all iterations, using a new reference image drastically reduces the accuracy of the network in both metrics. For all cases accuracy falls magnitudes of orders in both metrics. On the other hand, changing the reference image does not degrade quantitative performance on the registration by optimization based tools. We use same reference brain and test dataset of optimization based tools and find similar pattern as before therefore ANTS obtain best scores, Elastix and NiftyReg achieve subsequent positions.The accuracies of these traditional tools are much higher then the Voxelmorph. This experiment clearly shows a major drawback of deep-learning based Voxlemorph compared to its traditional counterparts.   

\begin{table}[t]
		\begin{center}
			\caption{Results on New Reference Brain}
			\label{Table:more_test}
			\begin{tabular}{l c c c c c}
				\hline
				Methods		& iteration & brain 1 & brain 1 & brain 2 & brain 2\\
						&  & CC & MI & CC & MI\\
				\hline
                ANTS		&100				&0.8226		&0.6778			&0.7294		&0.6338\\
				Elastix		&1000				&0.7891		&0.5818			&0.7591		&0.5246\\
                NiftyReg	&1000				&0.6149		&0.5530			&0.6544		&0.5722\\
                VM  &5000			&0.2900		&0.2037			&0.1944		&0.2025\\
				VM  &10000		&0.3161		&0.2221			&0.2255		&0.2243\\
				VM  &20000		&0.3161		&0.2221			&0.2255		&0.2243\\
				VM  &30000		&0.3649		&0.2255			&0.2613		&0.2034\\
                VM &40000		&0.3690		&0.2378			&0.2656		&0.2162\\
                VM  &49340		&0.3694		&0.2417			&0.2684		&0.2225\\
				\hline
			\end{tabular}
		\end{center}
	\end{table}
    
\section{Conclusion}
In this evaluation, we test and compare traditional optimization based registration algorithms with new deep-learning based approach on tissue cleared dataset. On this new and challenging dataset, traditional tools ANTS, Elastix and NiftyReg performed better than AIR and IRTK. Among the five traditional tools, ANTS obtained best quantitative and qualitative performance but failed to achieve best performance in efficiency on which Elastix outperformed all other traditional tools. Despite good accuracy all of these traditional tools are quite slow. The qualitative and quantitative performance of deep-learning based Voxelmorph registration method shows promising results comparable to ANTS, Elastix and NiftyReg. The quantitative accuracy of Voxelmorph is still behind the best performer ANTS. The gradual improvement of quantitative performance in both resolution level indicates its slow but steady learning characteristics. The great advantage of the method over traditional methods is its very high efficiency with 10s to 15s registration time. This promising performance indicates the potential of deep-learning based methods in image registration research. Despite having good qualitative and quantitative performance with excellent time-efficiency, the method has a number of serious drawbacks. Firstly, it requires a large dataset to train the network. Training on a large 3D dataset also requires heavy computational resources. Secondly, the method can operate only at 15\% resolution. These shortcomings clearly shows the requirement of further research and investigation on deep-learning based registration techniques to make them applicable in generic use.

\section*{Acknowledgments}
Computational resources and services used in this work were provided by the HPC and Research Support Group, Queensland University of Technology, Brisbane, Australia.

\bibliographystyle{ieeeconf}
\bibliography{new_refs.bib}
\end{document}